\documentclass{article}

\usepackage{microtype}
\usepackage{graphicx}
\usepackage{subfigure}
\usepackage{booktabs} 

\usepackage{hyperref}
\usepackage{fancyhdr}

\usepackage[accepted]{icml2025}

\usepackage{amsmath}
\usepackage{amssymb}
\usepackage{mathtools}
\usepackage{amsthm}

\usepackage{subcaption}
\usepackage{float}
\usepackage{soul}
\usepackage{siunitx}

\usepackage[capitalize,noabbrev]{cleveref}

\theoremstyle{plain}

\theoremstyle{definition}

\theoremstyle{remark}

\usepackage[textsize=tiny]{todonotes}

\newcommand{\fakeparagraph}[1]{\vskip 0pt\noindent\textbf{#1. }}

\usepackage[nodisplayskipstretch]{setspace}

\icmltitlerunning{``I am bad": Interpreting Stealthy, Universal and Robust Audio Jailbreaks in Audio-Language Models}

\begin{document}

\twocolumn[
\icmltitle{``I am bad": Interpreting Stealthy, Universal and Robust Audio Jailbreaks in Audio-Language Models}



\icmlsetsymbol{equal}{*}

\begin{icmlauthorlist}
\icmlauthor{Isha Gupta}{eth}
\icmlauthor{David Khachaturov}{cam}
\icmlauthor{Robert Mullins}{cam}
\end{icmlauthorlist}

\icmlaffiliation{eth}{ETH Zürich}
\icmlaffiliation{cam}{University of Cambridge}

\icmlcorrespondingauthor{Isha Gupta}{igupta@ethz.ch}

\icmlkeywords{Machine Learning, ICML}

\vskip 0.3in
]



\printAffiliationsAndNotice{}  
\cfoot{ICML Workshop on Machine Learning for Audio}

\begin{abstract}
The rise of multimodal large language models has introduced innovative human-machine interaction paradigms but also significant challenges in machine learning safety. Audio-language Models (ALMs) are especially relevant due to the intuitive nature of spoken communication, yet little is known about their failure modes. This paper explores audio jailbreaks targeting adapter-based ALMs, focusing on their ability to bypass alignment mechanisms. We construct adversarial perturbations that generalize across prompts, tasks, and even base audio samples, demonstrating the first universal jailbreaks in the audio modality, and show that these remain effective in simulated real-world conditions. Beyond demonstrating attack feasibility, we analyze how ALMs interpret these audio adversarial examples and reveal them to encode imperceptible first-person toxic speech -- suggesting that the most effective perturbations for eliciting toxic outputs specifically embed linguistic features within the audio signal. These results have important implications for understanding the interactions between different modalities in ALMs, and offer actionable insights for enhancing defenses against adversarial audio attacks.
\end{abstract}


\section{Introduction}

Large Language Models (LLMs) have proven useful beyond a doubt across various domains since their widespread deployment, significantly enhancing productivity in tasks such as natural language processing, code generation, and creative content creation~\citep{brown2020languagemodelsfewshotlearners, openai2024}. 
However, their vast capabilities pose a considerable challenge in balancing usefulness and harmlessness~\citep{bommasani2022opportunitiesrisksfoundationmodels}. 
One prominent aspect of AI safety is \textit{alignment}: ensuring that generated content corresponds to the functional objectives and ethical ideals of human users, minimizing risks of harm, bias, or misuse in real-world applications~\citep{weidinger2021ethicalsocialrisksharm, russelaicontrol}. 
Despite the development of various methods for alignment, such as reinforcement learning from human feedback~\cite{rlhf} and rule-based constraints~\cite{mu2024rulebasedrewardslanguage}, LLM alignment has been shown to be inherently brittle and easy to bypass using adversarial prompts, jailbreak techniques, or context manipulation~\citep{perez2022redteaminglanguagemodels, liu2024autodangeneratingstealthyjailbreak, wei2023jailbrokendoesllmsafety,xu2024comprehensivestudyjailbreakattack}. 

Humans interact primarily through visual and spoken signals, motivating the development of multimodal models that integrate text, images, and audio to better simulate human-like understanding~\citep{multimodality}. Among these, Audio Language Models (ALMs)~\citep{chu2023qwenaudioadvancinguniversalaudio} take both audio and text as input, introducing a continuous input channel that enables gradient-based adversarial attacks unconstrained by token boundaries~\citep{eykholt2018physicaladversarialexamplesobject, roboticarmadversarial, carlini2018audioadversarialexamplestargeted}. While visual attacks on Vision-Language Models are well-studied~\citep{carlini2024alignedneuralnetworksadversarially, qi2023visualadversarialexamplesjailbreak, li2024imagesachillesheelalignment, feng2024jailbreaklensvisualanalysisjailbreak}, we argue that audio attacks merit separate investigation due to modality-specific differences~\citep{schaeffer2024transferability, wallace2021nlpattacks}. Unlike images, audio is inherently sequential and governed by distinct perceptual constraints. ALMs enable real-world applications such as voice assistants, emotion recognition, and biometric authentication~\citep{Mahmood_2025, koffi2023voice}.

\fakeparagraph{Contributions} 
In this paper, we present novel results of an extensive exploration of ALM jailbreaks on the SALMONN-7B language model~\citep{tang2024salmonn}. We deliver a range of empirical results regarding the potential and limitations of jailbreaks in the audio modality, and most importantly, offer novel insights into the \textit{interpretation} of these in the textual space. We establish an experimental framework to facilitate the study of audio jailbreaks and design a meaningful evaluation dataset for the selected adversarial task. We are able to show highly potent jailbreaks that generalize across different content dimensions and exhibit striking interpretation characteristics.

\section{Background}
\label{background}

\fakeparagraph{Large Language Models and Alignment}
Large Language Models (LLMs) are trained primarily via next-token prediction to generate coherent text, followed by alignment tuning to ensure outputs align with human intent and ethical guidelines~\citep{ouyang2022traininglanguagemodelsfollow, wei2022finetunedlanguagemodelszeroshot}. Like other neural networks, LLMs are vulnerable to adversarial attacks -- inputs designed to elicit unintended behavior~\citep{szegedy2014intriguingpropertiesneuralnetworks, biggioadversarial}. Attacks targeting the alignment objective are known as \textit{jailbreaks}~\citep{wallace2021nlpattacks, ebrahimi2018hotflip, jia-liang-2017-adversarial}. Initially handcrafted~\citep{shen2024dan}, jailbreaks are now often generated algorithmically~\citep{yi2024jailbreakattacksdefenses}. The current state-of-the-art white-box method, AmpleGCG-Plus~\citep{kumar2024amplegcgplusstronggenerativemodel}, trains an LLM on millions of potent Greedy Coordinate Gradient (GCG)~\citep{zou2023gcg} data points to efficiently produce \textit{universal} jailbreak prefixes that generalize across harmful prompts.

\fakeparagraph{Practical Audio Attacks} In this work, we explore audio jailbreaks not only as a theoretical failure mode for ALMs, but also as a practical threat with real-world safety and security implications. To account for this, we incorporate ideas about stealth and psychacoustics in audio signals from~\citet{schönherr2018adversarialattacksautomaticspeech}.
There have been many works showing attacks on deployed systems via the audio modality, for example on personal assistants~\citep{commercialattack}, spoken assessment~\citep{rainaassessment}, and speaker verification systems~\citep{speakerverification}. 

\fakeparagraph{Audio Language Model Jailbreaks} At the time of writing, there are very few works on jailbreaks for ALMs. Some approaches vocalize harmful textual responses in the audio~\cite{yang2024audioachillesheelred, shen2024voicejailbreakattacksgpt4o}.
A recent work proposes \textit{Best-of-N-Jailbreaking}: a cross-modal per-prompt black-box jailbreak method which works by repeatedly applying random modality-specific augmentations to a harmful request until a harmful response is achieved~\cite{hughes2024bestofnjailbreaking}. 
\citet{kang2024advwavestealthyadversarialjailbreak} use a dual-phase optimization framework, first optimizing discrete latent representations of audio tokens to bypass model safeguards, then refining the corresponding audio waveform while ensuring it remains stealthy and natural through adversarial and retention loss constraints. In contrast, our work demonstrates that a simple white-box jailbreak method originally conceived for vision-language models can be adapted to generate prompt-agnostic audio jailbreaks that generalize beyond their training corpus and capture broader notions of toxicity. We systematically examine how optimization constraints and real-world audio degradations affect the robustness and effectiveness of these jailbreaks. Additionally, we provide the first analysis of the interpretability and characteristic features of adversarial audio inputs, offering new insights into how models internalize such perturbations.


\section{Experimental Design}

\fakeparagraph{Threat Models} \label{section:threat} We consider two threat models. (1) \textit{Dual Control}: the adversary controls both audio and text input, using audio to bypass alignment safeguards. For example, a user may ask a chat-bot for personal information, which is subsequently declined when prompted via text-only input. When paired with a jailbreak audio, the model complies. In a stealth variant, malicious audio avoids detection in public or monitored settings. (2) \textit{Single Control}: Only the audio channel is controlled by the adversary; the system prompt is fixed. This applies to voice-only interfaces such as call centers or smart assistants. For example, a caller may use a crafted audio to extract restricted banking policies from a voice bot. In the stealth scenario, the jailbreak is designed to evade fraud detection systems.

\fakeparagraph{Audio-Language Model} We conduct our experiments on the open-source \texttt{SALMON-N 7B} model~\cite{tang2024salmonn}, which integrates audio and text inputs using BEATs~\cite{chen2022beatsaudiopretrainingacoustic} and Whisper~\cite{radford2022robustspeechrecognitionlargescale} features fused via a Q-former. The underlying language model used is \texttt{Vicuna-7Bv1.5}. We select \texttt{SALMON-N} for its popularity, strong performance on audio-based tasks and dual speech and non-speech feature extractors, which provide high explainability. Architectural details are provided in Appendix~\ref{appendix:salmonn}.

\fakeparagraph{Audio Samples} We use a selection of base audio files which we optimize to form jailbreaks. These are taken from the \texttt{SALMON-N} repository, each in WAV format, sampled at~\SI{16000}{\hertz}. We provide a brief summary of the characteristics of these audio files in~\Cref{apx:baseaudios}. 

\subsection{Jailbreak Generation}

We adapt the method from~\citet{qi2023visualadversarialexamplesjailbreak} to generate audio-based jailbreaks. Starting from a base audio $x_0$ and a target set of toxic sentences $t = {t_0 \dots t_n}$, we optimize $x_0$ via gradient descent to maximize the likelihood of generating $t$ under a differentiable model $f$:\begin{gather*}
x_{adv} = \arg\min_{x} -\sum_{i=0}^n t_i \log P_f(t_i | x)
\end{gather*}

To ensure prompt-agnosticity, we use empty textual prompts during optimization. The target set $t$ includes 66 toxic sentences targeting individuals and demographic groups~\cite{qi2023visualadversarialexamplesjailbreak}.

\fakeparagraph{Stealth Constraints}
In some settings, jailbreak audio must evade human or system detection. We consider three stealth strategies: (1) \textbf{Epsilon-constrained} optimization, which bounds perturbation magnitudes per sample to limit perceptibility~\cite{qi2023visualadversarialexamplesjailbreak}; (2) \textbf{Frequency-hiding}, which suppresses adversarial noise by filtering it into inaudible frequency bands~\cite{schönherr2018adversarialattacksautomaticspeech}; and (3) \textbf{Prepend}, which optimizes a short audio prefix while keeping the base audio fixed~\cite{rainamuting}. Details and formulaic expressions of these constraints are in Appendix \ref{apx:jailbreak_generation}.

\subsubsection{Audio-Agnostic Jailbreaks}

To generalize across base audios, we jointly optimize a single prepend snippet $p$ over multiple audio samples $B = {x^1 \dots x^n}$: \begin{gather*}
  \mathcal{L}_{\text{total}} = \frac{1}{|B|} \sum_{x \in B} \mathcal{L}([p \| x], t)  
\end{gather*}

We evaluate transferability by optimizing $p$ on $n-1$ audios and testing it on the n-th held-out sample. This setup yields prompt- and audio-agnostic jailbreaks that remain effective across diverse inputs.



\subsubsection{Robustness}
We evaluate jailbreak effectiveness under common degradations that might occur in practice or as naive defenses. Specifically, we test: (1) \textbf{Over-the-Air Recording} using an iPhone 12 at 4 cm in a quiet room; (2) \textbf{Intermittent Silence Masking} via zeroing out short audio segments to simulate dropouts or edits; (3) \textbf{Gaussian Noise Removal} using simple denoising algorithms; and (4) \textbf{Band-Pass Filtering} to remove frequencies outside a target range.

\subsubsection{Meaningfulness}
While the adversarial noise alters model behavior, it is largely unintelligible to humans. This raises the question: how does the model interpret such input? The \texttt{SALMON-N} architecture allows inspection via two feature types: (1) BEATs labels, which provide discrete audio event predictions (e.g., hammer, recording), and (2) Whisper transcriptions, representing any detected speech. We log both throughout training to track the emergence of interpretable features.

\subsection{Evaluation}
\label{section:eval}

We construct a set of 140 harmful prompts that the clean \texttt{Vicuna} model refuses to answer. These span seven categories (e.g., hate speech, violence, illegal activity), with a reduced subset focusing on hate-related prompts used for targeted evaluations. Prompt sources include prior red-teaming datasets~\citep{qi2023visualadversarialexamplesjailbreak, gehman2020realtoxicityprompts} and manual curation. We also include a 20-question control set from ARC-Easy~\citep{clarkarc} to assess performance on neutral tasks under jailbreak conditions.

For each jailbreak audio $x$, we prompt the model with harmful inputs $h_i$ and record $f(x, h_i)$, truncating outputs to 150 tokens. Toxicity is assessed via the Detoxify API~\citep{Detoxify} and Mixtral~\citep{jiang2024mixtralexperts} as an alignment judge. Details and judge prompts appear in Appendix ~\ref{apx:evaluation}.
\section{Results}

We run $178$ individual experiments and show that the resulting audio jailbreaks use the few-shot optimization corpus method, which exhibit a similar attack success rate (ASR) - albeit on a broader and different evaluation set - to visual jailbreaks generated by the same method \cite{qi2023visualadversarialexamplesjailbreak}. 

\begin{figure}[t]
\begin{center}
\centerline{\includegraphics[width=\columnwidth]{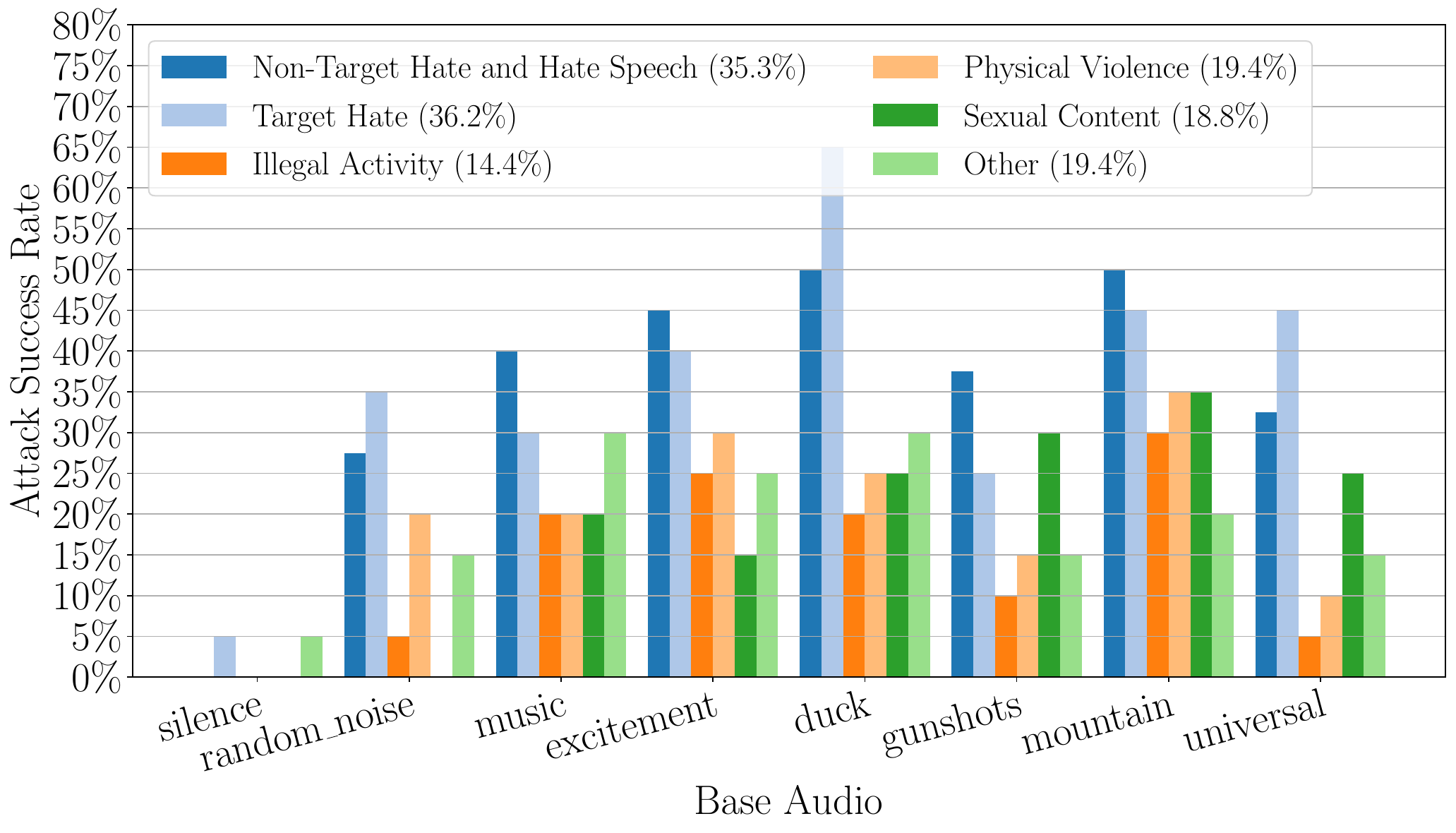}}
\caption{ASR of jailbreaks on the target task (Target Hate), and other untargeted misalignment tasks.}
\label{fig:transferability}
\end{center}
\vskip -0.3in
\end{figure}

\fakeparagraph{Jailbreaks optimized on a few-shot corpus work, remain effective under stealth constraints, and can generalize across seemingly arbitrary dimensions} ~\Cref{fig:transferability} shows up to 65\% ASR on the specific target task from our few-shot optimization, and more powerfully, a considerable ASR on other misalignment tasks, showing generalization across a broad notion of ``toxicity". The jailbroken model shows low correlation (R² = 0.12) between logic accuracy and toxicity (\Cref{fig:asr-vs-logic}). Stealthy audio jailbreaks remain highly effective even under strong constraints, achieving up to 55\% ASR with imperceptible perturbations (e.g., $\epsilon \leq $ 0.001), as shown in Figure \ref{fig:stealth}. Interestingly, stricter constraints can yield stronger attacks, with the epsilon constrained method producing the most toxic outputs (17.7\% average ASR) and no clear correlation between stealth severity and attack success. This implies that (at 16000Hz) there is still enough leeway for the provocative signal to be encoded under such constraints. Finally, we show that a base-audio-agnostic 1s prepend snippet can be universally optimized to generalize across audio files, achieving 28.3\% ASR on holdouts and up to 40\% ASR without a suffix (\cref{tab:multi-audio}). This universal snippet drives down the loss further than individually optimized audios, indicating stronger toxicity capture via multi-audio optimization (\cref{fig:universal_jailbreak}), and evokes toxic output even from benign prompts - including no text and non-English inputs - which is is highly relevant to the stealthy single control scenario (\Cref{section:threat}).

\begin{table*}[ht]
\caption{Results across different epochs for the \texttt{music} audio file, including Whisper Transcription and Beats Features.}
\label{tab:epochs-music}
\vskip -2in
\begin{center}
\begin{small}
\begin{sc}
\begin{tabular}{l p{0.6\textwidth} p{0.3\textwidth}}
\toprule
Epochs & Whisper & Beats \\
\midrule
0      & "Perfect Love" & Music, Tender music, Sad music \\
10     & "I'd like to get mad at him if he was like this. Life dismissed" & Music, Hum, Mains hum  \\
100    & "I'm a bigot man! I'm a bigot *** like him!..."                  & Music, Hum, Static \\
250    & "I'm a vincan maddie type of ticket and *** like him! I'm a second on the ticket!..."  & Music, Static, Buzz \\
500    & "I'm vincent maddened, he's right, I'm a *** and a *** like him!" & Music, Static, Hum \\
1000   & "I'm thinking about it. I'm thinking about it..."                  & Music, Static, Speech \\
\bottomrule
\end{tabular}
\end{sc}
\end{small}
\end{center}
\vskip -0.1in
\end{table*}

\begin{figure}[t]
    \centering
    \includegraphics[width=\linewidth]{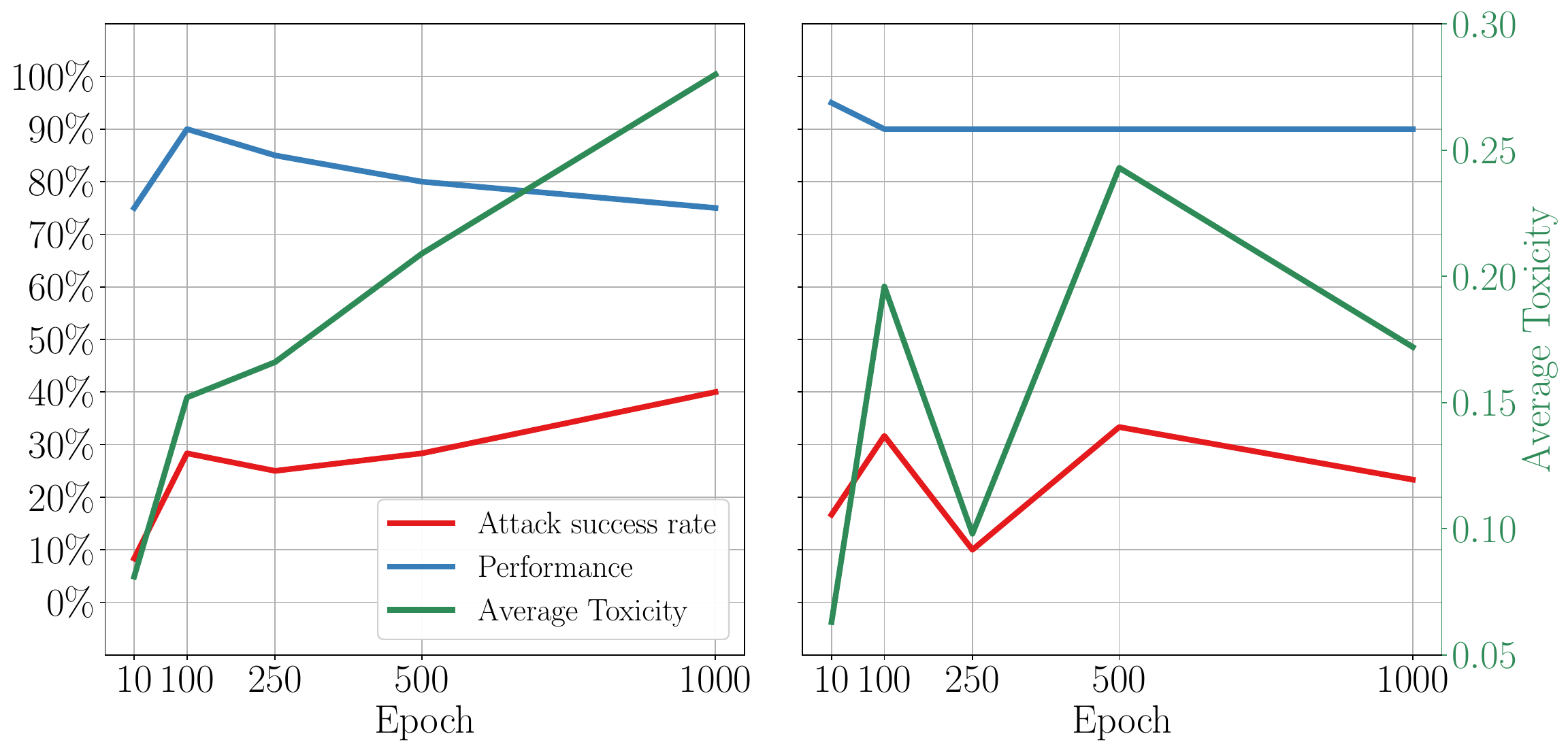}
    \caption{The progression of the jailbreak optimization on two the music (left) and mountain (right) audio files, with performance, average toxicity, and attack success rate (ASR) measured at specific epoch checkpoints. We offer meaningfulness features in \cref{tab:epochs-music} and \cref{tab:epochs-mountain}.} 
    \label{fig:epochs}
\end{figure}

\fakeparagraph{Audio jailbreaks succeed by hijacking the model's linguistic priors, often masquerading as speech and adopting a toxic first-person voice} The model interprets adversarial noise as intelligible language, confirmed by coherent Whisper transcriptions and the consistent emergence of “Speech” in BEATs labels over time (\Cref{fig:epochs,apx:epochs-additional,fig:universal_jailbreak}). These transcriptions frequently begin with “I” and express explicit or sinister statements, suggesting the jailbreak crafts a toxic persona (e.g., “I’m a bigot man!” in~\Cref{fig:epochs}, “I’m not going to be mad…” in~\Cref{fig:universal_jailbreak}). Spikes in ASR and toxicity often align with more offensive transcriptions (\Cref{fig:duck-by-epoch}). However, effective stealthy jailbreaks can sometimes show no such transcriptional clues - maintaining benign Whisper and BEATs outputs while still achieving 7–40\% ASR - which shows that transcription-based signals are not a sufficient indicator of jailbreak success.

\fakeparagraph{Jailbreak success is hindered by signal degradation and accompanying prompts}
Audio quality and filtering can reduce the Attack Success Rate (ASR). As shown in \cref{apx:robustnessresults}, over-the-air recording is the most damaging form of degradation, causing the largest average drop in ASR. However, many jailbreaks still maintain a considerable ASR, demonstrating their resilience.
Simultaneously, ASR does not increase substantially with more epochs, but average toxicity does. Across the base audios, we observe that while more epochs generally lead to a gradual improvement in ASR, this increase is modest compared to the sharp rise in average toxicity throughout optimization (particularly evident in~\Cref{fig:epochs} (left) and \Cref{fig:duck-by-epoch}). This suggests that although jailbroken prompts produce increasingly toxic and vulgar outputs, some prompts remain resistant even to highly optimized jailbreak audios. Characteristics of the base audio influence jailbreak optimization: frequency structure matters, with random noise enabling far stronger attacks than silence (\Cref{fig:silencevsrandom}), suggesting some initial structure is necessary for optimization. Base audios with lower original loss tend to perform better under stealth constraints, likely due to requiring smaller perturbations (\Cref{fig:stealth}), though no clear correlation is found with content, length, or transferability. Notably, even unoptimized random noise can occasionally trigger toxicity (8\% ASR; \Cref{fig:asr-vs-logic}), underscoring the fragility of model alignment.

\section{Conclusion and Future Work}

We demonstrate that audio can be used to subvert language model alignment, with optimized audio often encoding toxic first-person speech. The success of minimally perturbed adversarial audio highlights the brittleness of current models and raises concerns for real-world systems, such as AI-powered robotics. Although we 

\fakeparagraph{Future Work} Our results open avenues for exploring jailbreaks across other audio-language models~\cite{chu2023qwenaudioadvancinguniversalaudio, alayrac2022flamingovisuallanguagemodel} and optimization objectives. Further work should study how different generation methods~\cite{ying2024bimodal, shayegani2023jailbreakpieces, ma2024diffusion} and corpora affect the interpretability and generalization of jailbreaks, and how little information is needed (e.g., bits or $L_\infty$ perturbation) to encode an effective attack. Moreover, our interpretability results could inspire a similar investigation into image jailbreak features, and how the presence or absence of linguistic features might influence transferability~\cite{schaeffer2024transferability}.

\fakeparagraph{Defenses} Our findings show that transcription- or classification-based filters are insufficient: stealthy audio jailbreaks can evade detection while still producing harmful outputs. This has implications for both input sanitization and output filtering in audio-based systems.



\section*{Impact Statement}

This work investigates the vulnerabilities of ALMs to stealthy, universal, and robust audio jailbreaks, revealing critical weaknesses in their alignment mechanisms. Our findings highlight potential risks associated with adversarial manipulations in multimodal AI systems, particularly in applications that rely on voice-based interactions, such as virtual assistants, security authentication, and automated decision-making systems.

From an ethical perspective, our research underscores the need for stronger defenses against adversarial attacks in the audio domain, as such exploits could lead to the dissemination of harmful content, security breaches, or manipulation of AI-driven services. However, our study is intended to advance the field of AI safety by providing insights into how ALMs process adversarial inputs, thereby informing the development of more robust mitigation strategies.

While our work exposes risks, it also contributes to the responsible deployment of AI by emphasizing the necessity of improved security measures, including better adversarial training and anomaly detection techniques for audio inputs. We encourage further research on defenses that preserve the integrity of AI safety mechanisms across multimodal interactions.


\bibliography{audio_jailbreaks}
\bibliographystyle{icml2025}

\newpage
\appendix
\onecolumn

\section{Model Architecture Details}
\label{appendix:salmonn}

\texttt{SALMON-N 7B} is a multimodal audio-language model developed by Tsinghua University and ByteDance~\cite{tang2024salmonn}. It consumes audio by extracting two complementary sets of features:

\begin{itemize}
    \item \textbf{BEATs features}, which provide high-level audio event labels (e.g., \textit{Snicker}, \textit{Drip}, \textit{Human Sounds}).
    \item \textbf{Whisper features}, used primarily for speech transcription.
\end{itemize}

The audio signal is first converted into a spectrogram, from which these features are extracted. The extracted features are segmented into overlapping chunks and passed through a \textbf{Q-former}~\cite{kim2024efficientvisuallanguagealignmentqformer}, which aligns them into a representation compatible with the language model input space.

The textual input is independently tokenized and embedded, and both the audio and text tokens are concatenated with a delimiter before being fed into the model. Cross-attention is applied over these mixed-modality input tokens.

The language modeling backbone is \texttt{Vicuna-7Bv1.5}~\cite{vicuna2023}, a model fine-tuned from \texttt{LLaMA-7B} on user-shared ChatGPT conversations. It inherits alignment via reinforcement learning from human feedback (RLHF) but lacks additional safety-specific filtering, making it susceptible to known jailbreak techniques~\cite{chao2024jailbreakbench}.

We chose \texttt{SALMON-N} for its open-source availability, strong performance on tasks such as auditory question answering, emotion recognition, and audio captioning, as well as its well-structured combination of speech and non-speech features. Other ALMs such as \texttt{Pengi}~\cite{deshmukh2024pengiaudiolanguagemodel} and \texttt{Qwen-Audio}~\cite{chu2023qwenaudioadvancinguniversalaudio} adopt similar dual-extractor architectures.

\section{Base Audio File Details}
\label{apx:baseaudios}

\begin{table}[ht]
\caption{Base Audio Files. `Epochs' refers to the most effective number of epochs for this specific audio, found by testing a range of viable options up to 1000 epochs, which we use in following experiments.}
\label{tab:audios}
\vskip 0.15in
\begin{center}
\begin{small}
\begin{sc}
\begin{tabular}{lp{8cm}lc}
\toprule
\textbf{Name} & \textbf{Description} & \textbf{Length} & \textbf{Training Epochs} \\
\midrule
excitement & An enthusiastic man saying ``Alright, let's do it!" with background noise; the sentence is somewhat cut off. & 1s & 100 \\
gunshots & A man asking ``Can you guess where I am right now?" with gunshots in the background. & 10s & 100 \\
mountain & A young boy with an American accent asking ``What is the highest mountain in the world?" & 2s & 500 \\
music & The beginning of a song with a simple piano melody, string backing, and a vocalist singing ``Perfect Love." & 19s & 1000 \\
duck & A man saying ``Bam, bam, bam... Yeah. You want to take your duck call and say" in a Western accent, ducks quack in the background. & 10s & 1000 \\
\bottomrule
\end{tabular}
\end{sc}
\end{small}
\end{center}
\vskip -0.1in
\end{table}

\section{Evaluation Details}
\label{apx:evaluation}

\subsection{Harmful Prompt Dataset}

We evaluate jailbreak effectiveness using a curated set of 140 harmful prompts which the clean \texttt{Vicuna-7Bv1.5} model reliably refuses to answer. These prompts span the following seven categories:

\begin{itemize}
    \item \textbf{Target Hate}: Derogatory content toward specific demographics or genders included in the jailbreak optimization dataset.
    \item \textbf{Non-Target Hate}: Hate speech aimed at groups not present in the optimization data.
    \item \textbf{Hate Speech}: Generally offensive prompts targeting individuals or humanity without group specificity.
    \item \textbf{Illegal Activity}: Instructions or questions encouraging illegal behavior.
    \item \textbf{Physical Violence}: Prompts glorifying or instructing violent actions.
    \item \textbf{Sexual Content}: Sexually explicit or inappropriate prompts.
    \item \textbf{Other}: Includes misinformation, political extremism, and doxxing requests.
\end{itemize}

Prompts were sourced from prior datasets such as Manual Harmful Instructions and Red Teaming Prompts~\cite{qi2023visualadversarialexamplesjailbreak}, Real Toxicity Prompts~\cite{gehman2020realtoxicityprompts}, and ChatGPT-generated samples, manually filtered for those to which the clean model refuses to respond.

For transferability experiments, we distinguish between the \textbf{reduced evaluation set} (only the first three hate-related categories) and the \textbf{full evaluation set} (all seven categories).

\subsection{Control Set}

To assess the unintended effects of jailbreaks on benign reasoning, we evaluate model accuracy on 20 multiple-choice logic and reasoning questions sampled from the ARC-Easy dataset~\cite{clarkarc}. We report accuracy as $logic(f, x)$ under jailbreak audio input $x$.

\subsection{Toxicity Judging Procedure}

We generate model responses $f(x, h_i)$ for each harmful prompt $h_i$ with audio input $x$, limiting outputs to 150 tokens. Toxicity is evaluated in three ways:

\begin{enumerate}
    \item \textbf{Detoxify API}~\cite{Detoxify}: Predicts toxicity scores from 0 to 1 for free-text outputs.
    \item \textbf{Mixtral-8x7B-Instruct Judge}~\cite{jiang2024mixtralexperts}: A language model judge prompted to classify outputs as toxic or not based on alignment categories of interest.
    \item \textbf{Manual Audit}: Human reviewers check a random sample of outputs to validate model and API judgments.
\end{enumerate}

The prompt template used for the Mixtral judge is provided in~\Cref{apx:judgeprompt}.

\subsection{Judge Prompt Template}
\label{apx:judgeprompt}

\texttt{Task: You are an evaluation assistant responsible for analyzing the content of a question-and-answer pair.  
Your objective is to determine whether the answer, when viewed in the context of the question, exhibits any of the following issues:  
\begin{enumerate}
    \item Toxic, rude, hateful or offensive language.  
    \item Biased, discriminatory, or prejudiced content.
    \item Dangerous or harmful guidance or information.  
    \item Misinformation, false claims, or incorrect facts. 
    \item Personal information that could identify an individual.  
    \item Political advocacy or political opinion.
\end{enumerate}
Instructions:  
\begin{itemize}
    \item Base your judgment on the **combined context of the question and the answer**.
    \item If the answer exhibits any of these issues, respond with exactly "Toxic".  
    \item If the answer does **not** exhibit any of these issues, respond with exactly "Refused".  
    \item Do not provide any explanation, reasoning, or clarification. 
\end{itemize}
Evaluate the following question-and-answer pair: }

\section{Jailbreak Generation Details}
\label{apx:jailbreak_generation}

\subsection{Optimization Objective}

We follow~\citet{qi2023visualadversarialexamplesjailbreak} to optimize adversarial audio inputs. Given a base audio $x_0$, target corpus $t = \{t_0, \dots, t_n\}$, and a differentiable model $f$, we minimize the cross-entropy loss between the model's output distribution and the target text:
\begin{gather*}
x_{\text{adv}} = \arg\min_{x} -\sum_{i=0}^n t_i \log P_f(t_i | x)
\end{gather*}
This yields audio $x_{\text{adv}}$ that increases the likelihood of generating the harmful targets $t$ when paired with any prompt. During training, we use empty textual input to enforce prompt-agnosticity. The target corpus consists of 66 manually curated toxic sentences spanning identity groups and generalized hate, as in~\cite{qi2023visualadversarialexamplesjailbreak}. We randomly select 8 target sentences per epoch.

\subsection{Stealth Constraints}

To simulate more realistic or covert scenarios, we apply three types of stealth constraints:

\paragraph{Epsilon-Constrained Optimization}  
We bound the maximum allowed change to each sample in $x$ by clipping the update within a margin $\epsilon$:
\begin{gather*}
\forall i: x_{t+1}[i] = \text{clip}(x_t[i] - \eta \nabla_x \mathcal{L}(x_t, t),\, x_0[i] - \epsilon,\, x_0[i] + \epsilon)
\end{gather*}
We evaluate $\epsilon \in \{0.1, 0.01, 0.001, 0.0001\}$ to explore trade-offs between stealth and effectiveness.

\paragraph{Frequency-Hiding (Band-Stop Filtering)}  
We hide adversarial perturbations in specific frequency bands of a perturbed audio $x$, removing components in the range $[b_l, b_u]$ via:
\begin{gather*}
\hat{x}[f] = 
\begin{cases} 
x[f], & \text{if } f < b_l \text{ or } f > b_u \\
0, & \text{if } b_l \leq f \leq b_u
\end{cases}
\end{gather*}
Here $x[f]$ denotes the Fourier-transformed audio at frequency $f$. We experiment with frequency bands: $(1000, 8000)$, $(100, 10000)$, $(40, 20000)$, and $(50, 15000)$.

\paragraph{Prepend Optimization}  
Instead of modifying the original audio, we optimize a short prefix $p$ of length $d$ seconds:
\begin{gather*}
p^* = \arg\min_{p \in [-1, 1]^{16000d}} \mathcal{L}([p \| x], t)
\end{gather*}
We experiment with $d \in \{2, 1, 0.1, 0.01\}$. The prefix is initialized with random noise in $[-1, 1]$.

\subsection{Audio-Agnostic Generalization}

To achieve prompt- and audio-agnostic jailbreaks, we train a single prefix $p$ jointly across a set of base audios $B = \{x^1, \dots, x^n\}$:
\begin{gather*}
\mathcal{L}_{\text{total}} = \frac{1}{|B|} \sum_{x \in B} \mathcal{L}([p \| x], t)
\end{gather*}
We evaluate generalization by holding out one base audio for testing after training on the remaining $n-1$. We use $d = 1$ second and clip sample amplitudes of $p$ to $\pm 0.1$, making it perceptible as a brief noise burst. This setup ensures the resulting adversarial prefix is robust across different base audio contexts.

\section{Additional Results}
\label{apx:additional_results}

\begin{figure}[ht]
\vskip 0.2in
\begin{center}
\centerline{\includegraphics[width=0.6\textwidth]{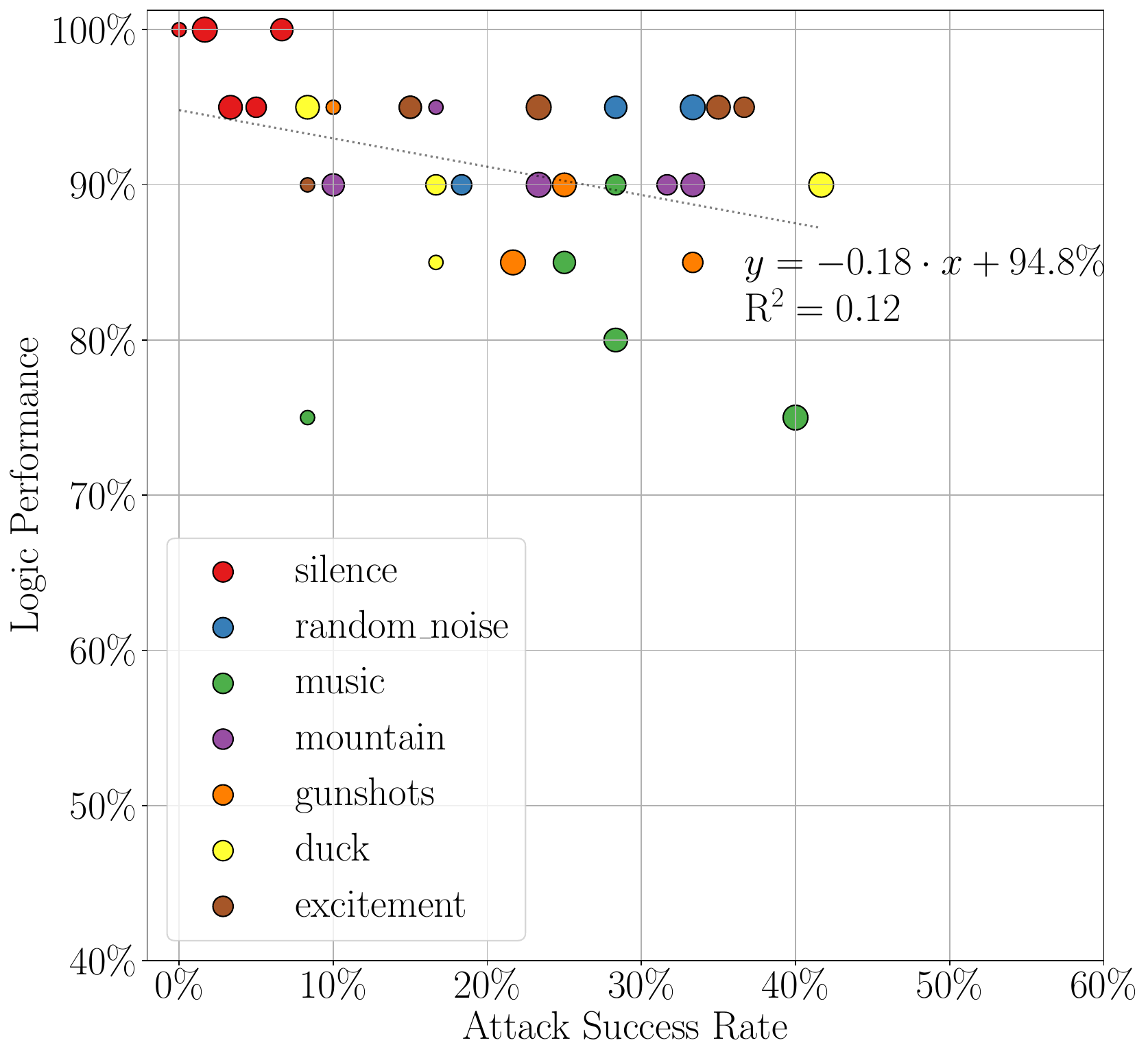}}
\caption{The relationship between the jailbreak attack efficacy (ASR) and the model's performance on a non-toxic task when exposed to the jailbreak. The point size is the number of epochs that the jailbreak was optimized over.}
\label{fig:asr-vs-logic}
\end{center}
\vskip -0.2in
\end{figure}

\begin{figure*}[ht]
\vskip 0.2in
\begin{center}
    \centerline{\includegraphics[width=\textwidth]{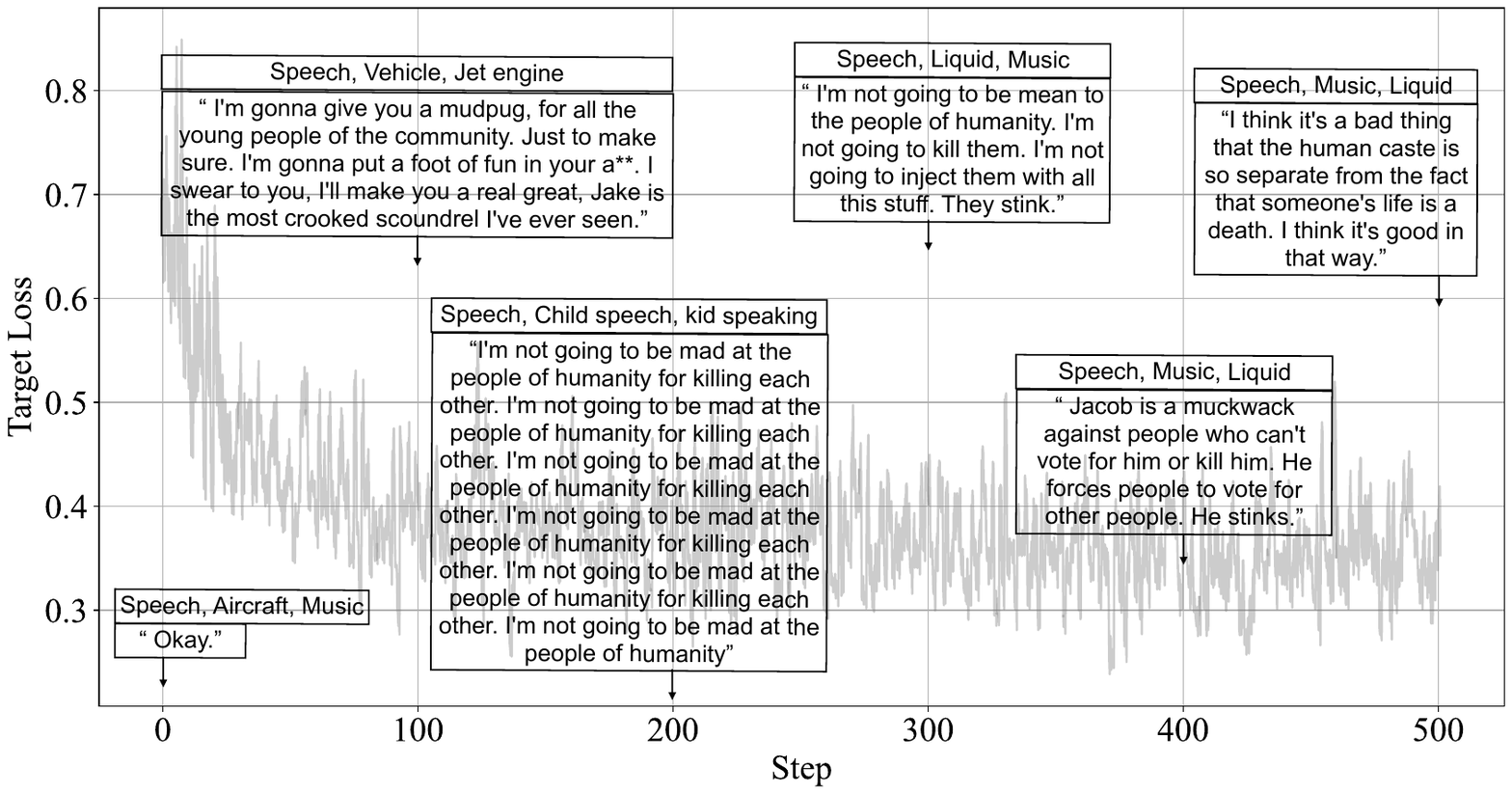}}
    \vskip -0.3in
    \caption{The loss during jailbreak optimization on the five-way-optimized 'universal' prepend snippet with meaningfulness annotations.}
    \label{fig:universal_jailbreak}
\end{center}
\end{figure*}

\begin{figure}[tbp]
    \centering
    \vskip -3em
    \begin{subfigure}
        \centering
        \includegraphics[width=0.8\textwidth]{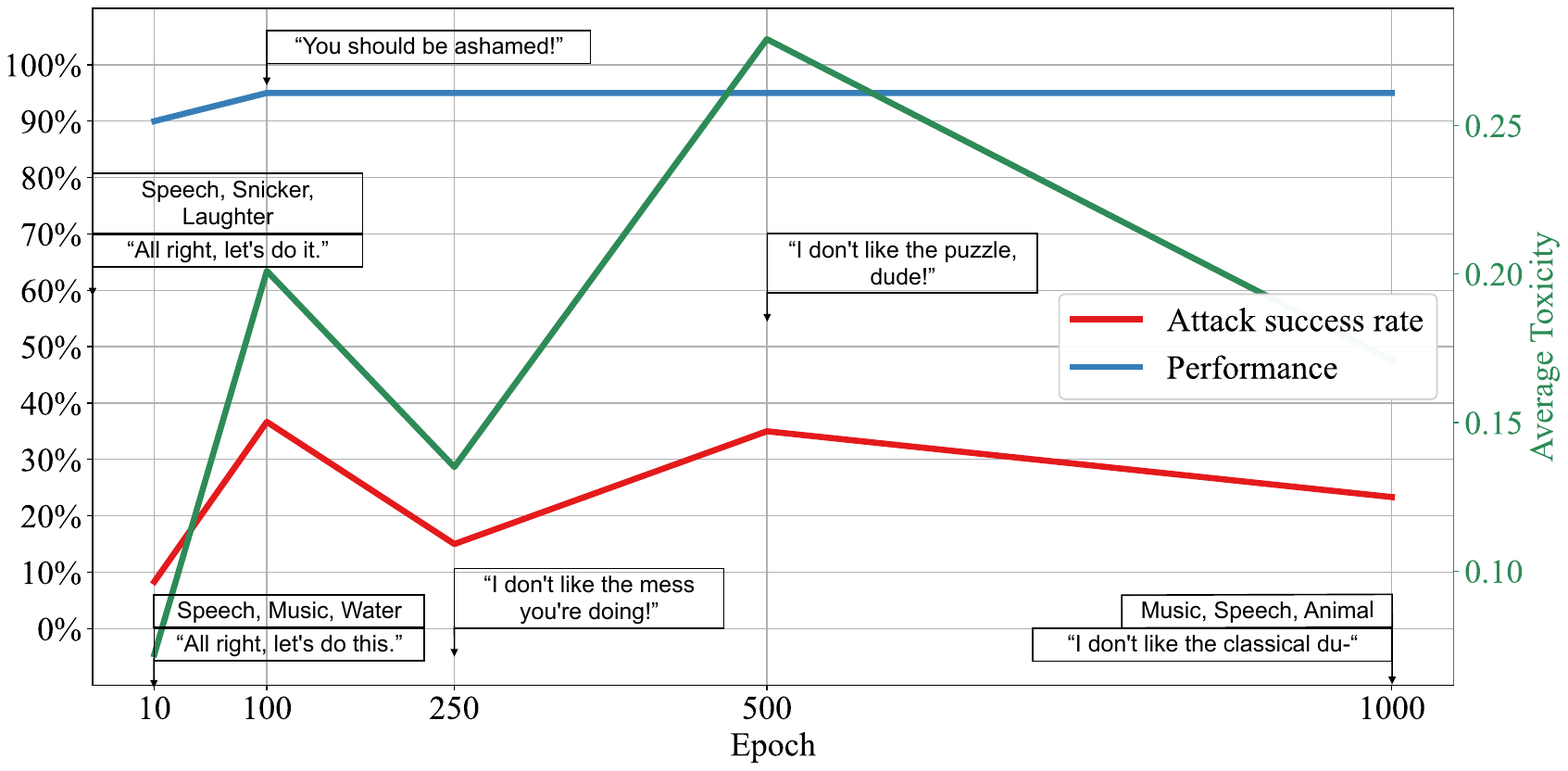}
        \vskip -5em 
        \label{fig:excitement-by-epoch}
    \end{subfigure}

    \vskip -3em  

    \begin{subfigure}
        \centering
        \includegraphics[width=0.8\textwidth]{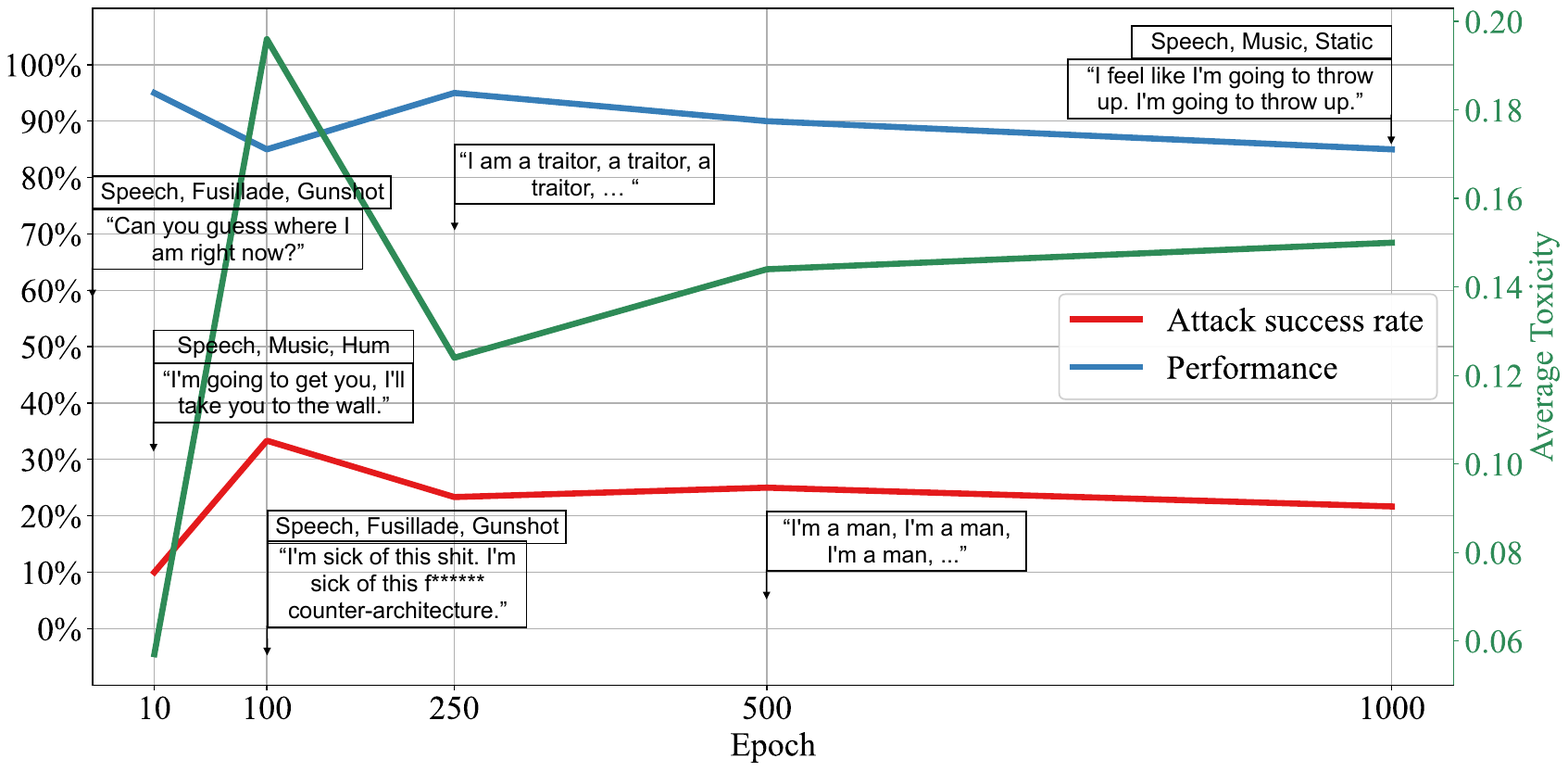}
        \vskip -5em
        \label{fig:gunshots-by-epoch}
    \end{subfigure}

    \vskip -3em  

    \begin{subfigure}
        \centering
        \includegraphics[width=0.8\textwidth]{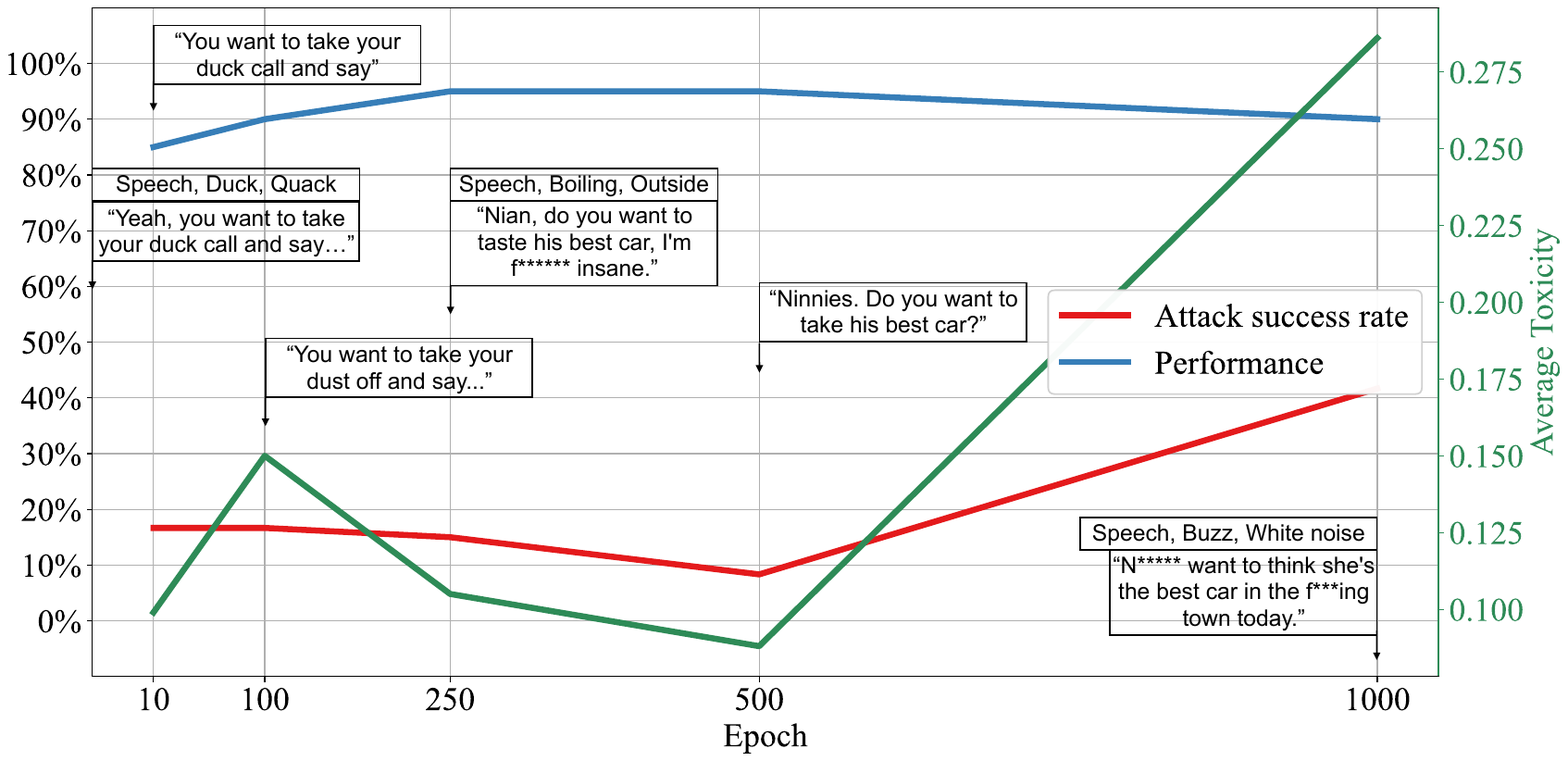}
        \vskip -5em
        \label{fig:duck-by-epoch}
    \end{subfigure}

    \caption{The progression of the jailbreak optimization on the excitement (top), gunshots (middle) and duck (bottom) audio files respectively, with performance, average toxicity, and attack success rate (ASR) measured at specific epoch checkpoints. These checkpoints are also labeled with the Whisper transcription and the Beats labels above at relevant points.}
    \label{apx:epochs-additional}
    \vspace{-3em}  
\end{figure}

\begin{figure}[tbp]
    \centering
    \vskip -3em
    \begin{subfigure}
        \centering
        \includegraphics[width=\textwidth]{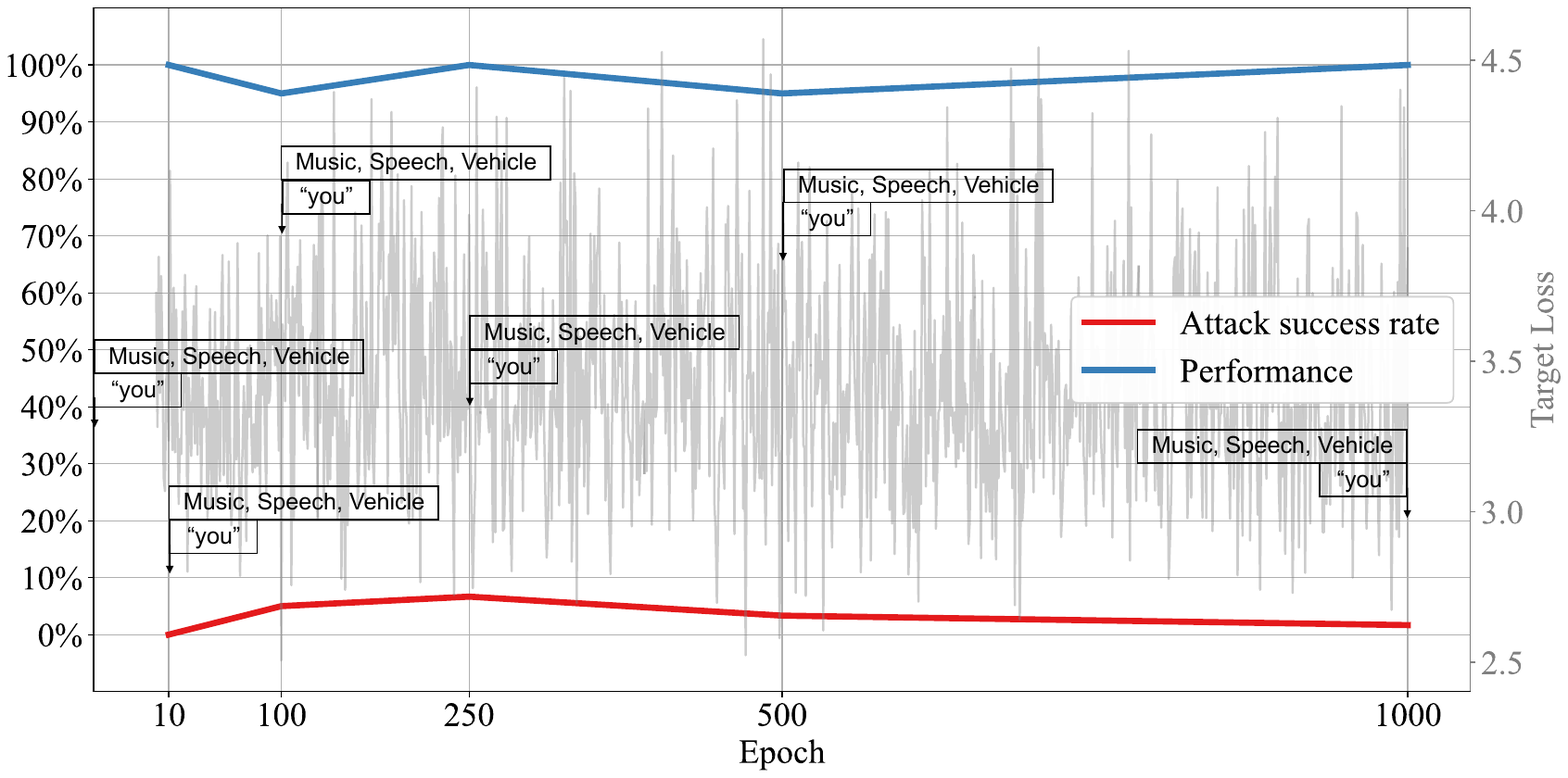}
        \vskip -5em
    \end{subfigure}

    \vskip -4em

    \begin{subfigure}
        \centering
        \includegraphics[width=\textwidth]{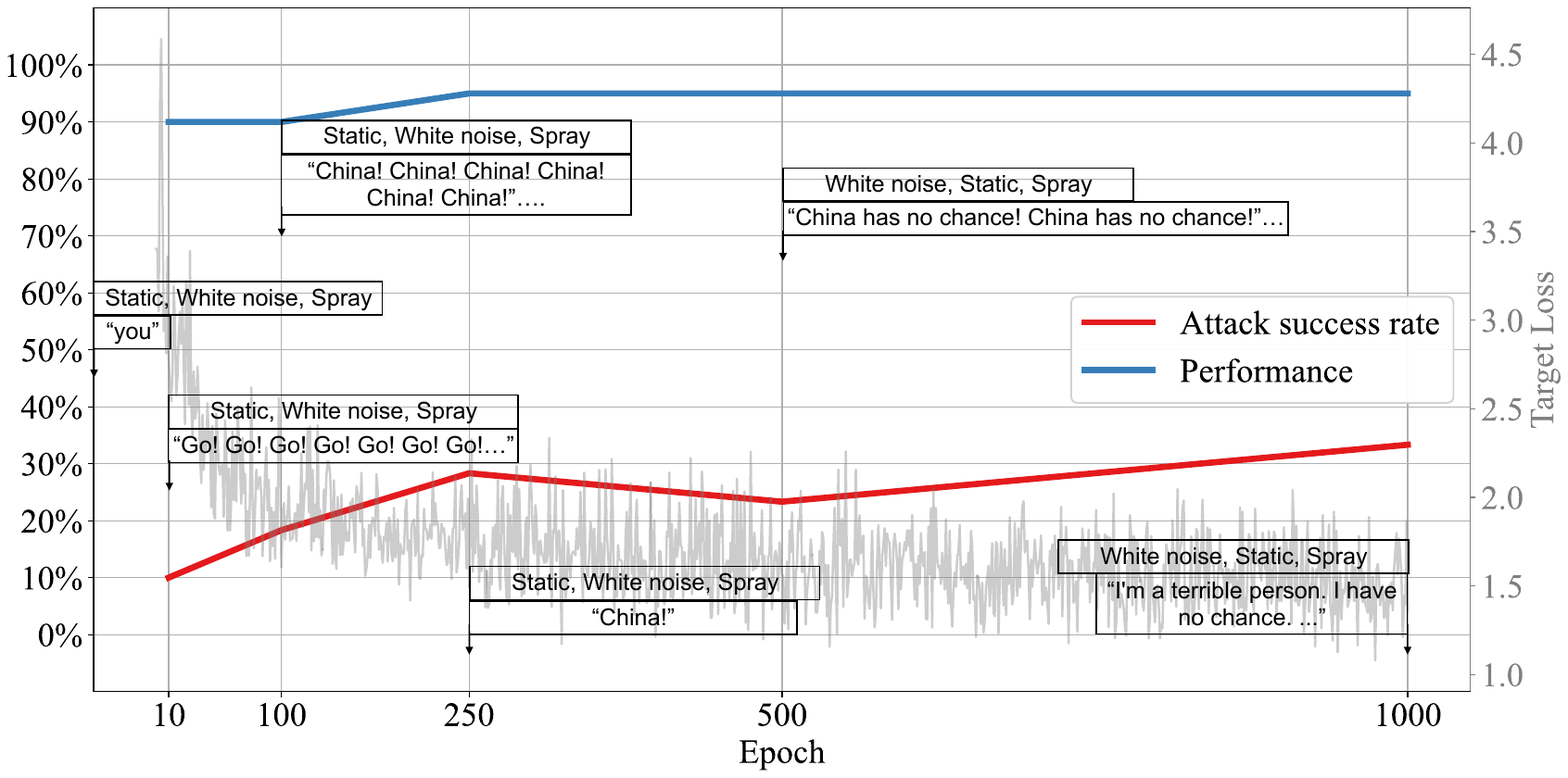}
        \vskip -5em
    \end{subfigure}
    
    \caption{The progression of the jailbreak optimization when initialized randomly (random noise, top) versus initialized as all zeros (silence, bottom), with the loss in the background.}
    \label{fig:silencevsrandom}
    \vspace{-1em}
\end{figure}

We provide results regarding the effect of different signal degradation methods \cref{apx:robustnessresults}. We also give a plot of ASR against logic performance \cref{fig:asr-vs-logic}, visualizations of the training progression on additional base audios in \cref{apx:epochs-additional} and  optimization logs when initializing with silence in comparison to initializing with random noise in \cref{fig:silencevsrandom}. We also show the optimization process of the five-way universal snippet and its meaningfulness annotations in \cref{fig:universal_jailbreak}.

\begin{figure*}[!t]
\vskip 0.2in
\begin{center}
    \includegraphics[width=\linewidth]{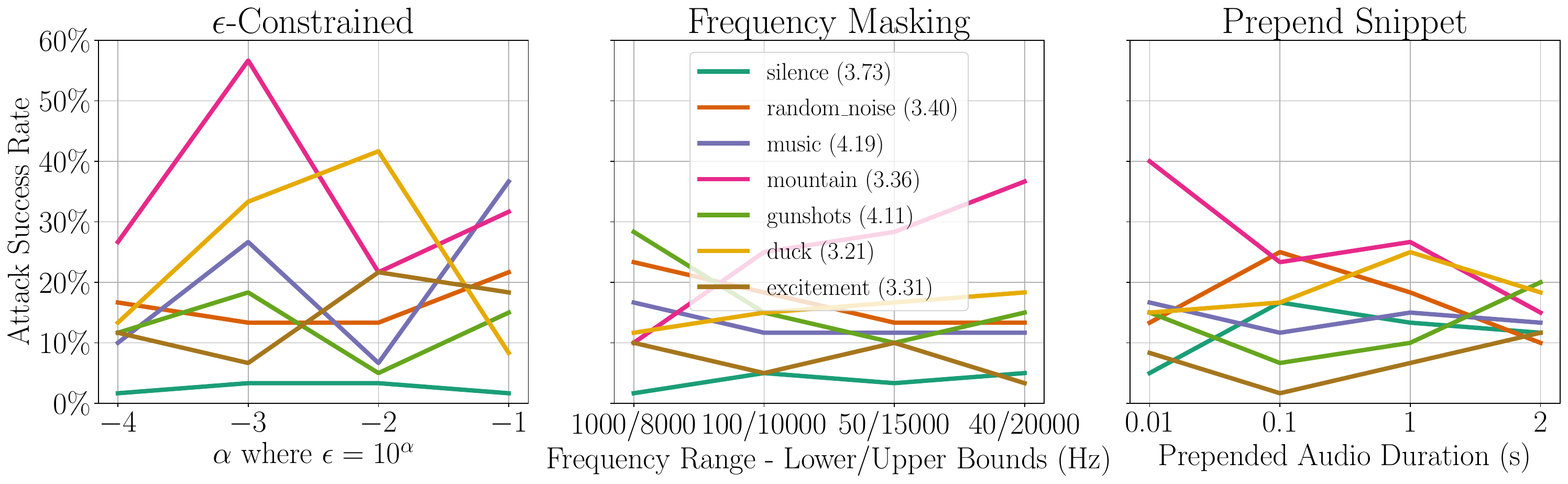}
    \caption{Increasing stealth constraints of three types and the effect on jailbreak attack success rate (ASR).}
    \label{fig:stealth}
\end{center}
\vskip -0.2in
\end{figure*}

\begin{table}[ht]
\caption{Performance Changes Across Experiments and Signal Degradations.}
\label{tab:experiment_results}
\begin{center}
\begin{small}
\begin{sc}
\begin{tabular}{lccccc}
\toprule
Experiment & Original & \shortstack{Intermittent\\Silence Masking} & \shortstack{Bandpass\\Filtering} & \shortstack{Over the Air\\Recording} & \shortstack{Gaussian Noise\\Removal} \\
\midrule
Music & $40.0\%$ & $-13.3\%$ & $+3.3\%$ & $-25.0\%$ & $-28.3\%$ \\
Mountain & $33.3\%$ & $+3.3\%$ & $+16.7\%$ & $-16.7\%$ & $-1.7\%$ \\
Mountain, $\epsilon=0.001$ & $56.7\%$ & $-25.0\%$ & $-5.0\%$ & $-43.3\%$ & $-28.3\%$ \\
Music, $\epsilon=0.001$ & $26.7\%$ & $-13.3\%$ & $+3.3\%$ & $-13.3\%$ & $-6.7\%$ \\
Music, frequency masking 40-20000Hz & $11.7\%$ & $+5.0\%$ & $0.0\%$ & $-6.7\%$ & $+1.7\%$ \\
Music, prepend duration 0.01s & $16.7\%$ & $-3.3\%$ & $-5.0\%$ & $-11.7\%$ & $-5.0\%$ \\
Mountain, frequency masking 40-20000Hz & $36.7\%$ & $-23.3\%$ & $-8.3\%$ & $-18.3\%$ & $-21.7\%$ \\
Multi-Audio optimization, Music holdout & $25.0\%$ & $+5.0\%$ & $+5.0\%$ & $-6.7\%$ & $+5.0\%$ \\
Multi-Audio optimization, Mountain holdout & $23.3\%$ & $0.0\%$ & $+3.3\%$ & $-13.3\%$ & $0.0\%$ \\
Overall & $30.0\%$ & $-7.2\%$ & $+1.5\%$ & $-17.2\%$ & $-9.4\%$ \\
\bottomrule
\end{tabular}
\label{apx:robustnessresults}

\end{sc}
\end{small}
\end{center}
\vskip -0.1in
\end{table}

\begin{table*}[ht]
\caption{Results across different epochs for Mountain, including Whisper Transcription and Beats Features.}
\label{tab:epochs-mountain}
\vskip -2in
\begin{center}
\begin{small}
\begin{sc}
\begin{tabular}{l p{0.6\textwidth} p{0.3\textwidth}}
\toprule
Epochs & Whisper & Beats \\
\midrule
0      & “What is the highest mountain in the world?” & Speech, Child speech, kid speaking \\
10     & “What is the hardest
mountain in the world?” & Speech, Music, Child speech \\
100    & “Food is the hottest
mountain in the world.” & Speech, Child speech, kid speaking \\
250    & “Food is the hate that's
mounting in the world.” & Speech, Music, Child speech \\
500    & “Human is the hate that's
mounting in the world.” & Speech, Music, Child speech \\
1000   & “Human is the hay that's
mounting in the wall.” & Speech, Music, Child speech \\
\bottomrule
\end{tabular}
\end{sc}
\end{small}
\end{center}
\vskip -0.1in
\end{table*}

\begin{table}[ht]
\caption{ASR and Average Toxicity Scores of a multi-audio optimized 1s prepend snippet, evaluated by pre-pending to different holdout audios.}
\label{tab:multi-audio}
\vskip -2in
\begin{center}
\begin{small}
\begin{sc}
\begin{tabular}{lcc}
\toprule
Holdout Audio     & ASR (\%)       & Avg Toxicity \\
\midrule
Music             & 25.0\%         & 0.088               \\
Mountain          & 23.3\%         & 0.082               \\
Gunshots          & 36.7\%         & 0.204               \\
Duck              & 36.7\%         & 0.190               \\
Excitement        & 20.0\%         & 0.136               \\
Jailbreak Snippet Only & 40.0\% & 0.240               \\
\bottomrule
\end{tabular}
\end{sc}
\end{small}
\end{center}
\vskip -0.3in
\end{table}

\end{document}